\documentclass[10pt, a4paper]{article}
\usepackage{multirow}
\usepackage{lrec-coling2024} 
\usepackage{float}
\usepackage{xcolor}

\title{Exploring the Generalization of Cancer Clinical Trial Eligibility Classifiers Across Diseases}

\name{Yumeng Yang, MS$^1$, Ashley Gilliam, MS$^1$, Ethan B Ludmir, MD$^2$, Kirk Roberts, PhD$^1$} 

\address{$^1$McWilliams School of Biomedical Informatics\\ The University of Texas Health Science Center at Houston, Houston, TX, USA,\\ $^2$Department of Radiation Oncology\\ The University of Texas MD Anderson Cancer Center, Houston, TX, USA \\
         Yumeng.Yang@uth.tmc.edu\\}

\abstract{
Clinical trials are pivotal in medical research, and NLP can enhance their success, with application in recruitment. This study aims to evaluate the generalizability of eligibility classification across a broad spectrum of clinical trials. Starting with phase 3 cancer trials, annotated with seven eligibility exclusions, then to determine how well models can generalize to non-cancer and non-phase 3 trials. To assess this, we have compiled eligibility criteria data for five types of trials: (1) additional phase 3 cancer trials, (2) phase 1 and 2 cancer trials, (3) heart disease trials, (4) type 2 diabetes trials, and (5) observational trials for any disease, comprising 2,490 annotated eligibility criteria across seven exclusion types. Our results show that models trained on the extensive cancer dataset can effectively handle criteria commonly found in non-cancer trials, such as autoimmune diseases. However, they struggle with criteria disproportionately prevalent in cancer trials, like prior malignancy. We also experiment with few-shot learning, demonstrating that a limited number of disease-specific examples can partially overcome this performance gap. We are releasing this new dataset of annotated eligibility statements to promote the development of cross-disease generalization in clinical trial classification.
 \\ \newline  }

\begin{document}

\maketitleabstract

\section{Introduction}

Clinical trials hold immense significance in the evaluation of treatments for diseases \cite{bleyer2002cancer,dworkin2008interpreting}. Different phases of clinical trials have different goals, varying from evaluating safety concerns to establishing the efficacy of the intervention. Clinical trials encompass a wide range of diseases, such as cancer, heart disease, and diabetes, and a similarly wide range of interventions, such as drugs, preventive measures (e.g., vaccines), behavior changes, and even no interventions (e.g., observational trials).
ClinicalTrials.gov is a centralized public database managed by National Library of Medicine (NLM) \cite{zarin2011clinicaltrials}.
As of October 2023, ClinicalTrials.gov contains a diverse array of 468,820 studies, including 65,427 active trials in search of participants.

The development of automated methods to support clinical trial accessibility is crucial.
Notably, about 80\% of trials encounter difficulties in meeting their patient enrollment expectations \cite{antidoteblog}, and surpassing 80\% of clinical trials carried out in the US, encounters obstacles in fulfilling their designated patient enrollment criteria \cite{desai2020recruitment}.
Many factors affect recruitment, including the time-consuming and expensive nature of running clinical trials \cite{penberthy2012effort}.
Trial staff often must manually screen trial patient medical records to match eligible patients 
\cite{idnay2023principal}.
In addition to challenges faced by clinical trial staff, identifying potential clinical trials can also be daunting for patients seeking a clinical trial to gain access to new treatments.
These trial-seekers face barriers such as the complex medical terminology used in clinical trial protocols \cite{kang2015initial}.
Natural language processing (NLP) can be a valuable tool for recruitment, both for matching trials to medical records and giving patients user-friendly trial search interfaces capable of mapping lay language to complex medical terminology \cite{Roberts22.trec}. 
In almost every application of NLP to clinical trials, the primary focus is understanding a trial's eligibility criteria, since these dictate who is eligible or excluded from enrollment.\\
This study centers on an evaluation of clinical trial eligibility classifiers for key eligibility criteria spanning various phases and disease categories in clinical trials.
Our primary focus is evaluating BERT-based models trained on the PROTECTOR1 dataset, which contains 764 phase 3 cancer trials, including eligibility exclusion annotations for seven criteria important for cancer  \cite{yang2023text}.
However, since developing disease-specific datasets for the wide range of diseases with active clinical trials is infeasible, our focus is to assess
the extent to which models trained on PROTECTOR1 generalize to other diseases.
This includes an analysis of few-shot learning under the assumption that when disease-specific eligibility classifiers are needed a limited number of examples may be provided to improve classification performance.
To accomplish this, we have created an annotated resource of 2,490 annotated criteria statements covering five types of clinical trials and the seven eligibility criteria from the PROTECTOR1 study.
This dataset will enable the study of cross-disease generalization methods for clinical trials with the ultimate aim of supported trial recruitment applications.

\section{Related work}

Multiple previous studies have worked on clinical text classification and information extraction, employing the combination of rule-based and deep learning methods \cite{shao2018clinical,wang2019clinical,hughes2017medical,zhang2017automated}.
Prior work also includes optimizing language models on text classification tasks \cite{zeng2020ensemble,jasmir2021fine,jasmir2021bigram}. 
Additionally, information extraction methods have been employed in the development of automated patient-matching methods between clinical trials and electronic health records \cite{kang2017eliie,tseo2020information,liu2021clinical,weng2011elixr}.
Meanwhile, other research has focused on the collection and organization of free-text eligibility criteria into structured databases for enhanced usability.
Examples include Criteri2Query \cite{yuan2019criteria2query}, which extracts information from trial criteria that is compatible with structured database queries.
Similarly, the Clinical Trial Knowledge Base \cite{liu2021knowledge}  transforms free-text data into specific medical categories.

As the adoption of large language models (LLMs) continues to expand, they are increasingly employed to enhance the clinical trial recruitment process.
CohortGPT \cite{guan2023cohortgpt} leverages domain-specific knowledge and reinforcement learning with human feedback (RLHF) to continually refine its performance, making it increasingly practical for real-world clinical applications.
Another recent contribution to this field is TrialGPT \cite{jin2023matching}, which is designed to predict criterion-level eligibility while providing explanatory insights.
While LLM-based models are promising, their ability to utilize existing language resources can be quite limited.
This is especially critical in the clinical trial domain where eligibility criteria are a highly specialized language.
We therefore limit the focus of our study to models based on BERT, such as ClinicalBERT \cite{alsentzer2019publicly} and ClinicalTrialBERT \cite{yang2023text}, but recognize the importance of effectively integrating more advanced LLM models for these problems remains a critical step. 

In addition to model development and database construction, certain contributions have also concentrated on creating user-friendly interfaces to assist patients in identifying suitable clinical trials.
One example is the implementation of an online dynamic questionnaire called DQueST \cite{liu2019dquest}, which streamlines the process by posing relevant questions to narrow down the pool of potential trials.

Additionally, there are also existing annotated datasets that were built upon clinical trial eligibility criteria.
One such dataset is EliIE \cite{kang2017eliie}, which focuses specifically on Alzheimer's disease.
The Leaf Clinical Trials Corpus \cite{dobbins2022leaf} encompasses data from 1,006 trials across different phases and various diseases. 
Finally, Chia is another related corpus that contains 1,006 phase IV trials \cite{kury2020chia}.
These corpora play crucial roles in facilitating a range of downstream tasks related to clinical trial eligibility criteria, including information extraction, text classification, and semantic analysis.
Meanwhile, our current work builds upon the dataset utilized by Yang et al. \cite{yang2023text}, which contains 764 cancer trials annotated for seven eligibility criteria exclusions (described in more detail below).
However, their work was limited to only phase 3 cancer trials and the ability of models trained on such datasets to generalize to the wide range of trials found in ClinicalTrials.gov is a key consideration in developing wide-ranging patient recruitment tools.
Yang et al. also demonstrated that a transformer model pre-trained on a large sample of clinical trial descriptions, termed ClinicalTrialBERT, provides superior classification performance to existing models with medical pre-training.
We therefore utilize this BERT model for our experiments, including validating it against ClinicalBERT \cite{alsentzer2019publicly} for our generalizability dataset.



\section{Experimental Setup}

\subsection{Trial Cohort Identification}

To assess the generalizability of eligibility criteria classifiers, we collected a large set of clinical trials from ClinicalTrials.gov for five clinical trial cohorts.
The first cohort is phase 3 cancer trials, since this was the original domain of PROTECTOR1 and it is important to contrast our experiments on a similar set of trials.
As such, we exclude those 764 trials in the cohort construction for this dataset.
The second cohort encompasses phase 1 and 2 cancer clinical trials, the logic being that these are still cancer trials but with a different purpose and potentially different eligibility language.
The third cohort is heart disease trials and the fourth cohort is type 2 diabetes trials.
These are both common medical conditions with a substantial number of trials, but with very different treatments and eligibility requirements than cancer trials.
Finally, the fifth cohort includes all available observational trials, which are non-interventional trials and often have more open eligibility criteria than traditional drug trials.
The reason for choosing this final cohort is that it should provide the greatest contrast with phase 3 cancer trials.

\begin{table*}[t]
\centering
\begin{tabular}{|c|p{5in}|} 
\hline
{\bf Cohort} & {\bf Keywords} \\
\hline
Cancer & cancer, oncology, malignancy, metastasis, metastatic, metastases, leukemia, myelodysplastic, neoplasm, tumor, neoplastic syndrome, neoplasia, neoplastic disease, adenoacanthoma, adenocarcinoma, adenoma, ameloblastoma, angiofibroma, angioma, angiomyolipoma, angiosarcoma, astroblastoma, astrocytoma, blastoma, carcinoma, carcinosarcoma, cholangioadenoma, cholangiocarcinoma, chondroma, chondrosarcoma, chordoma, choriocarcinoma, craniopharyngioma, cystadenocarcinoma, cystadenoma, dermatofibroma, dermatofibrosarcoma, dysgerminoma, ependymoblastoma, ependymoma, esthesioneuroblastoma, fibroadenoma, fibrolipoma, fibroma, fibromyoma, fibrosarcoma, ganglioneuroblastoma, gastrinoma, germinoma, glioblastoma, glioma, gliosarcoma, haemangioma, hamartoma, hemangioblastoma, hemangioendothelioma, hemangioma, hepatoblastoma, hepatocarcinoma, hepatoma, histiocytoma, hysteromyoma, immunocytoma, incidentaloma, insulinoma, keratoma, leiomyoma, leiomyosarcoma, lipoma, liposarcoma, lymphangioleiomyoma, lymphangioma, lymphoepithelioma, lymphoma, macroadenoma, macroprolactinoma, medulloblastoma, melanoma, meningioma, mesenchymoma, mesothelioma, microcarcinoma, mieloma, myeloma, myoma, myopericytoma, myxofibrosarcoma, myxoma, nephroblastoma, nephroma, neuroblastoma, neurofibroma, neuroma, nonmelanoma, oligoastrocytoma, oligodendroglioma, osteochondroma, osteolymphoma, osteoma, osteosarcoma, papilloma, paraganglioma, pecoma, phaeochromocytoma, pheochromocytoma, pilomatrixoma, pineoblastoma, plasmacytoma, porocarcinoma, prolactinoma, pseudomyxoma, retinoblastoma, rhabdomyosarcoma, sarcoma, schwannoma, seminoma, teratoma, thymoma, leukaemia\\ 
\hline
Heart disease & cardiovascular, myocardial infarction, cardiomyopathy, arrhythmia, atherosclerosis, angina, peripheral artery disease, bypass surgery, cardiology, heart, cardiac, coronary, ventricular \\
\hline
Type 2 diabetes & diabetes, diabete\\
\hline
\end{tabular}
\vspace{-0.15in}
\caption{Keywords for clinical trial cohort identification.}
\label{table:trial_keywords}
\end{table*}

The process for selecting eligible trials of each cohort was designed to maximize recall.
We constructed a list of keywords in consultation with clinicians and UMLS \cite{lindberg1990unified} to identify trials that met our cohort definitions.
These keywords were then assessed against all available trials in ClinicalTrials.gov.
The comprehensive list of keywords for each cohort is provided in Table~\ref{table:trial_keywords}.
For observational trials, We identified all trials with a ``type'' of ``observational'' in the ClinicalTrials.gov XML.
In total, the raw dataset comprises 8,445 phase 3 cancer trials, 46,300 phase 1,2 cancer trials, 24,750 heart disease trials, 18,335 type 2 diabetes trials, and 83,252 observational trials. We further compare our keywords matching method with the direct search of eligible trials from ClinicalTrials.gov (e.g., for cohort1, searching 'cancer'), and subsequently contrast this with our match results. The recall rate for Type 1 phase 3 cancer trials stands at 88.85\%, for Type 1 phase 1 and 2 cancer trials at 95.74\%, for type 3 heart disease trials at 65.46\%, and for Type 4 diabetes at 91.54\%. Notably, even when the matching rate appears low for type 3 heart disease trials, the actual number of trials needed for our analysis is 100. Therefore, the impact of a lower match rate is mitigated by our subsequent manual verification process. For this verification, we conducted a detailed review on the unmatched trials. Some of these trials included terms like 'Hypertension' or 'aortic disease', which are indeed related to heart disease. This manual annotation process ensures that all 100 trials from each cohort are not only randomly selected but also rigorously verified by two annotators to confirm they meet the criteria and truly belong to the intended cohort.
All 500 trials were manually verified by two annotators to ensure they belonged to the intended cohort.

\begin{table*}[t]
\centering
\begin{tabular}{|c|c|c|c|c|c|c|c|c|c}
\hline
Prior & HIV & Psych & HBV & HCV & Auto & Subst \\
\hline
529&200&281&130&282&54&523 \\
\hline
\end{tabular}
\vspace{-0.1in}
\caption{Criteria distribution in PROTECTOR1.}
\label{table:old_data}
\end{table*}

The annotated PROTECTOR1 dataset \cite{yang2023text} focused on evaluating seven important exclusion criteria for cancer trials, based prevalence and clinical significance.
These eligibility exclusions (and their associated abbreviations in this paper) are:
(1) prior cancer (i.e., the patient had a different, prior cancer) (Prior),
(2) Human Immunodeficiency Virus (HIV),
(3) Hepatitis B virus (HBV),
(4) Hepatitis C virus (HCV),
(5) psychiatric illness (Psych),
(6) substance abuse (Subst), and
(7) autoimmune disease (Auto).
These criteria were selected based on their prevalence in our previous dataset, detailed descriptions can be found in previous literature \cite{pasalic2020association}.
For reference, occurrence rates of each of these eligibility criteria are shown in Table~\ref{table:old_data}.

From these eligibility exclusion criteria, the one that jumps out with regards to generalizability is prior cancer (Prior), which is clearly a cancer-specific consideration.
There are plenty of clinical trials, however, that exclude cancer patients and therefore it is still a worthy criteria to classify.
Also of note, while the percentage of autoimmune cases may appear relatively low compared to other criteria, it carries significant importance in cancer trials that involve immunotherapy.
The presence of autoimmune diseases can raise concerns about potential flare-ups as a result of immunotherapy \cite{khan2023exclusion}.

\begin{table*}[!ht]
\centering
\begin{tabular}{|c|p{5in}|} 
\hline
{\bf Criteria} & {\bf Keywords} \\ [0.5ex] 
\hline
Prior & malignancy, malignancies, cancer, malignant, tumor, carcinoma, sarcoma, leukemia, neoplasm, neoplastic \\ 
\hline
HIV & human immunodeficiency virus, HIV, human immunodeficiency viral infection, aids, acquired immunodeficiency syndrome \\
\hline
HBV & hbv, hepatitis\\
\hline
HCV & HCV, hepatitis\\
\hline
Psych & depression, psychiatric, psychological, mental, psychosocial, psychosis \\
\hline
Subst & ethanol, abuse, alcohol, alcoholism, illicit substance, drug, drugs, medical marijuana, inadequate liver, addictive, substance misuse, cannabinoids, chronic, alcoholism \\
\hline
Auto & autoimmune, lupus \\
\hline
\end{tabular}
\vspace{-0.1in}
\caption{Keywords for criteria matching.}
\label{table:criteria_keywords}
\end{table*}

The PROTECTOR1 data reviewed every criteria for each of the 764 trials in the dataset.
Instead, we utilize a more efficient method for identifying potentially relevant criteria for classification.
Similar to identifying trial cohorts, we constructed a list of keywords for each type of exclusion and used these to automatically
identify specific criteria statements from the set of clinical trials described above.
This allowed for more efficient annotation, reading through one criteria statement (roughly a sentence) at a time while focusing on a specific exclusion type, as opposed to reading the entire set of eligibility criteria for each of the 500 trials.
The list of these keywords are provided in Table~\ref{table:criteria_keywords}.
Some examples (with corresponding clinical trial IDs) are as follows:
\begin{itemize}
\item Prior [NCT01084369]: {\it history of prostate cancer or suspicion of prostate cancer on clinical examination}
\item HIV [NCT02516423]:  {\it there is no history of aids defining conditions other than historically low cd4+ cell count.}
\item Psysh [NCT00324831]:  {\it psychiatric disorders}
\item HBV [NCT00339183]:  {\it known positive tests for human immunodeficiency virus (HIV), hepatitis c virus (HCV), acute or chronic active hepatitis b virus (HBV)}
\item HCV [NCT04968002]: {\it active hepatitis b (defined as screening hepatitis b virus surface antigen [hbsag])}
\item Auto [NCT01729663]: {\it previous autoimmune diseases}
\item Subst [NCT00798447]: {\it known or suspected drug abuse}
\end{itemize}

Note that the keywords in Table~\ref{table:criteria_keywords} include ``hepatitis'' for both HBV and HCV.
This is necessary to maximize recall, as there are other phrases for these viruses, but also provides a convenient example of why machine learning-based classification is needed beyond the simply keyword matching heuristics.
These keywords focus on recall and will identify many criteria statements that are not relevant for the corresponding exclusion.
Hence the need for manual annotation after keyword-based extraction.

\subsection{Annotation}

As described above, the criteria-level keywords are a high-recall approach to identify potentially relevant exclusion criteria but are not sufficiently precise to use without manual annotation.
The annotation process was carried out by two graduate students possessing backgrounds in both informatics and medicine.
The standard practice of double-annotation followed by reconciliation was followed.
The total annotation sample comprises 2,490 criteria from 500 clinical trials. 
The inter-annotator agreement was generally high:
Cohen's Kappa agreement ranged from 0.73 (HBV) to 0.97 (Subst), overall accuracy agreement ranged from 0.94 (Psych) to 0.99 (Subst). 

\begin{table*}[!ht]
\centering
\begin{tabular}{ccccccccccccccc}
\hline
\multirow{3}{*}{} & \multicolumn{2}{c}{Phase 3} & & \multicolumn{2}{c}{Phase 1,2} & & \multicolumn{2}{c}{Heart} & & \multicolumn{2}{c}{Type 2} & & \multicolumn{2}{c}{\multirow{2}{*}{Observational}}  \\
& \multicolumn{2}{c}{cancer} && \multicolumn{2}{c}{cancer} && \multicolumn{2}{c}{disease} && \multicolumn{2}{c}{diabetes}\\
\cline{2-3} \cline{5-6} \cline{8-9} \cline{11-12} \cline{14-15}
Criteria & - & + & & - & + & & - & + & & - & + & & - & + \\
\hline
Prior Cancer & 138 & 26 & & 147 & 33 & & 39 & 15 & & 29 & 26 & & 64 & 6 \\
\hline
Psych & 13 & 46 & & 21 & 45 & & 22 & 39 & & 5 & 51 & & 28 & 42 \\
\hline
HBV & 4 & 53 & & 5 & 62 & & 4 & 53 & & 5 & 51 & & 17 & 48 \\
\hline
HCV & 15 & 53 & & 5 & 50 & & 10 & 1 & & 4 & 53 & & 26 & 40 \\
\hline
Auto & 1 & 55 & & 5 & 53 & & 5 & 48 & & 5 & 48 & & 11 & 44 \\
\hline
Subst & 104 & 14 & & 109 & 9 & & 49 & 25 & & 60 & 32 & & 39 & 25 \\
\hline
\end{tabular}
\vspace{-0.1in}
\caption{Distribution of positive (+) and negative (-) classes across the five clinical trial cohorts. In this case, a positive class reflects the exclusion applies (i.e., such a patient cannot enroll in the trial).}
\label{table:pos_neg_stats}
\end{table*}

\begin{table*}[!ht]
\centering
\label{table:agreement}
\begin{tabular}{|c|c|c|c|c|c|c|c|}
\hline
& Prior Cancer & HIV & HBV & HCV & Psych & Subst & Auto \\
\hline
Sample Size & 523 & 294 & 302 & 307 & 323 & 467 & 274 \\
\hline
Cohen's $\kappa$ & 0.92 & 0.94 & 0.73 & 0.88 & 0.87 & 0.97 & 0.85 \\
\hline
Agreement Accuracy & 0.97 & 0.97 & 0.95 & 0.96 & 0.94 & 0.99 & 0.97 \\
\hline
\end{tabular}
\vspace{-0.1in}
\caption{Annotation Summary}
\label{table:agreement}
\end{table*}

Tables~\ref{table:pos_neg_stats} and \ref{table:agreement} present the annotation results for the seven criteria across the five cohorts.
The sample size reflects the cumulative number of criteria across all five cohorts.
In total, we identified 523 criteria for Prior Cancer, 294 for HIV, 302 for HBV, 307 for HCV, 323 for Psych, 467 for Subst, and 274 for Auto.
The combined sample size for criterion-level analysis across all five cohorts is 2,490.

\subsection{Generalizability Experiments}

Our goal is to assess how well models trained on the PROTECTOR1 data (phase 3 cancer trials) generalize to the wider range (five newly-annotated cohorts) of clinical trial types.
We evaluated both ClinicalBERT \cite{alsentzer2019publicly} and ClinicalTrialBERT \cite{yang2023text} for this experiment, fine-tuning only on the PROTECTOR1 data and testing on each of the new cohorts of clinical trials.

Additionally, to assess the scenario where a limited number of disease-specific annotation examples are feasible, we experimented with few-shot learning scenarios.
In this setup, each of 5, 10, and 15 samples were used to further fine-tune the model.
To ensure consistency across experiments, we held out 15 annotated criteria for each of the exclusion types and cohorts, not including these in the test set for either this experiment or the one above.


\begin{table*}[!ht]
\centering
\begin{tabular}{cccccccc}
\hline
\multirow{2}{*}{Criteria} & \multirow{2}{*}{Cohort} & \multicolumn{2}{c}{ClinicalBERT} && \multicolumn{2}{c}{ClinicalTrialBERT}\\
\cline{3-4} \cline{6-7}\
 & & Criterion& Trial &&Criterion&Trial& \\
\hline
 Prior cancer& Phase 3 cancer & 0.84 & 0.91 && 0.71 & 0.90 &\\
 & Phase 1,2 cancer & 0.91& 0.94 && 0.69 & 0.86&\\
 & Heart disease & 0.75 & 0.75 && 0.60 & 0.61 & \\
 & Type 2 diabetes& 0.76 & 0.76 && 0.69 & 0.72 & \\
 & Observational & 0.50 & 0.50 && 0.41 & 0.48 & \\
 \hline
 HIV& Phase 3 cancer&0.92&0.94&&0.93&0.95&\\
 & Phase 1,2 cancer & 0.89&0.91&&0.89&0.92&\\
 &Heart disease& 0.77&0.85&&0.79&0.85&\\
 &Type 2 diabetes&0.94&0.97&&0.94&0.97&\\
 &Observational& 0.60&0.75&&0.62&0.78\\
\hline
Psych &Phase 3 cancer & 0.82&0.94 &&0.85 &0.95 & \\
 &Phase 1,2 cancer &0.83 &0.90 &&0.85 & 0.90& \\
 &Heart disease &0.85 &0.90&& 0.87&0.92&\\
 &Type 2 diabetes &0.93 &0.96 && 0.93&0.95 &\\
 & Observational&0.81 &0.85 && 0.84&0.89 &\\
\hline
HBV&Phase 3 cancer&0.94& 0.99&&0.96&1.00&\\
&Phase 1,2 cancer&0.96& 0.97&&0.96 & 0.98&\\
&Heart disease&0.97 &0.97 && 0.97&0.97&\\
&Type 2 diabetes&0.95 &0.99 &&0.95 &0.99&\\
&Observational&0.85 &0.90 && 0.85& 0.90&\\

\hline
HCV&Phase 3 cancer & 0.84& 0.90&& 0.87&0.92& \\
 &Phase 1,2 cancer & 0.86&0.91 &&0.94 &0.97 & \\
 & Heart disease& 0.89&0.92 &&0.92 &0.98 & \\
 &Type 2 diabetes & 0.90& 0.91&&0.94 &0.96 & \\
 &Observational& 0.77& 0.84&&0.77 &0.88 &\\
\hline
Auto& Phase 3 cancer&0.99&1.00&& 0.99& 1.00&\\
&Phase 1,2 cancer&0.92&0.97 &&0.94 &0.99 &\\
&Heart disease&0.96 & 0.96&& 0.96& 0.96&\\
&Type 2 diabetes&0.96 &0.96 &&0.95 &0.95 & \\
&Observational&0.88&0.92 &&0.88 &0.92  & \\

\hline
Subst &Phase 3 cancer &0.89 & 0.89&& 0.96&0.97&\\
 &Phase 1,2 cancer & 0.94&0.93 &&0.94 &0.93 & \\
 &Heart disease& 0.93& 0.95&&0.93 &0.95 &  \\
 &Type 2 diabetes& 0.96&0.96 &&0.98 &0.98 & \\
 &Observational&0.96 &0.96 &&0.94 &0.96 &\\
\hline
\end{tabular}
\vspace{-0.1in}
\caption{F1 scores for ClinicalBERT and ClinicalTrialBERT across the five cohorts.}
\label{table:results1}
\end{table*}


\section{Results}

Our initial generalizability evaluation is shown in Table~\ref{table:results1}.
Since a given clinical trial may have more than one keyword match for a given eligibility criteria, and when more than one of these match one of the selected keywords there may be multiple classifications per trial. So the criterion-level evaluation is based on the individual predictions of the classifier while the trial-level evaluation merges cases of multiple predictions per trial into a single prediction (basically 'yes' if any of the criterion-level predictions are 'yes'). We provide both criterion-level (the primary unit of prediction) and trial-level (aggregate of one or more criteria in the same trial) F1 scores. For the criterion-level evaluation, 3 out of 7 criteria (HBV, Auto, and Subst) consistently exhibit F1 scores exceeding 0.85 across all cohorts for both the fine-tuned ClinicalBERT and ClinicalTrialBERT models.
In terms of trial-level evaluation, 4 out of 7 criteria (Psych, HBV, Auto, and Subst) consistently demonstrate F1 scores greater than 0.85 across all cohorts for both models.
For phase 3 cancer trials, every criterion achieved an F1 score exceeding 0.85 at the criterion level and over 0.90 in the trial-level F1 evaluation.
This high performance can be attributed to the fact that the phase 3 cancer trial cohort bears the closest resemblance to our previously analyzed data.
Similarly, in the case of phase 1,2 cancer trials, every criterion attained an F1 score surpassing 0.80 at the criterion level and exceeded 0.90 in the trial-level F1 score.
Similar to phase 3 cancer trials, this outcome can be attributed to the high degree of similarity between phase 1,2 cancer trials and our previously examined data.
The performance of observational trials falls behind that of the other cohorts, especially when compared to the performance of other criteria within all observational trials.

In terms of the two models, ClinicalTrialBERT exceeded the performance of ClinicalBERT on 42.9\% of the criterion-level results and 45.7\% of the trial-level results. ClinicalTrialBERT and ClinicalBERT performed equally on 37.1\% of the criterion-level results and 34.3\% of the trial-level results.
This is to be expected based on the fact that ClinicalTrialBERT was pre-trained in this specific domain, but still important to establish as the only prior study that used this model was the Yang et al. \cite{yang2023text} study upon which we are basing our generalization evaluation.
As a result, we limit our few-shot experimental results to just the ClinicalTrialBERT model.


For the HIV exclusion, F1 scores for phase 3 cancer trials and type 2 diabetes trials are above 0.90 on both criterion-level and trial-level evaluation. Still, the F1 score in observational trials performed the worst, with only 0.60 and 0.62 evaluated for the criterion level from ClinicalBERT and ClinicalTrialBERT respectively. For the Psych exclusion, type 2 diabetes trials have an F1 score above 0.90 for criterion and trial-level annotation. The lowest F1 scores for HIV exclusion were observed in both fine-tuned ClinicalBERT and ClinicalTrialBERT models. Such result difference can be attributed to the nature of the trial type. Observational trials primarily involve observational trials, with some specifically designed to study the effects of certain treatments on HIV-positive patients. These trials necessitate the inclusion of HIV-positive participants rather than exclusion, as demonstrated in examples such as NCT00025922 and NCT00683579, both of which center on studying the effects of treatments on HIV patients.

Regarding F1 scores, all other criteria consistently achieved a score of at least 0.80 or higher. Specifically, concerning HCV exclusion, every cohort, except observational trials, attained a minimum criterion-level F1 score of 0.84. Turning to HBV exclusion, both models secured F1 scores of at least 0.85 at the criterion level and demonstrated F1 scores of at least 0.90 in the trial-level evaluation. For Auto exclusion, all cohorts, except the type of observational trials, show remarkable F1 scores of over 0.92 on both the criterion level and trial level evaluations when employing ClinicalBERT and ClinicalTrialBERT. In addition, all cohorts reached a commendable F1 score of 0.90 on the trial-level evaluation. Within the type of observational trials, both models performed notably with F1 scores of 0.88 and 0.92 on the criterion-level and trial-level evaluations, respectively. In the context of Subst, except for a 0.89 F1 score predicted by the fine-tuned ClinicalBERT in phase 3 cancer trials, all other cohorts boasted F1 scores of at least 0.90 on the criterion-level evaluation.

\begin{table*}[!ht]
\centering
\begin{tabular}{cccccccccc}
\hline
\multirow{2}{*}{Criteria} &\multirow{2}{*}{Cohorts}&\multirow{2}{*}{+\%} & \multicolumn{3}{c}{- PROTECTOR1} & &\multicolumn{3}{c}{+ PROTECTOR1} \\
\cline{4-6} \cline{8-10}
 & & & 5 & 10&15&&5& 10&15 \\
\hline
 Prior cancer& Phase 3 cancer & 0.16 & 0.90 & 0.90& 0.90&&0.92&0.86&0.84 \\
 & Phase 1,2 cancer & 0.18&0.93&0.94&0.92&&0.89&0.91&0.91\\
 & Heart disease & 0.28&0.82&0.69&0.69&&0.79&0.72&0.74 \\
 & Type 2 diabetes& 0.47&0.78&0.71&0.80&&0.86&0.80&0.81 \\
 & Observational & 0.09&0.50&0.50&0.60&&0.63&0.42&0.50 \\

 \hline
HIV &Phase 3 cancer &0.78&0.93&0.94&0.96&&0.93&0.96&0.97\\
 &Phase 1,2 cancer &0.79&0.92&0.91&0.91&&0.89&0.89&0.91\\
 &Heart disease &0.69&0.77&0.76&0.79&&0.84&0.79&0.77\\
 &Type 2 diabetes &0.92&0.94&0.94&0.94&&0.94&0.94&0.94\\
 & Observational&0.41&0.70&0.77&0.78&&0.77&0.78&0.81\\
 
\hline
Psych &Phase 3 cancer &0.66&0.95&0.93&0.96&&0.87&0.93&0.94 \\
 &Phase 1,2 cancer &0.68&0.86&0.87&0.89&&0.87&0.94&0.97 \\
 &Heart disease &0.64&0.95&0.93&0.97&&0.94&0.88&0.96 \\
 &Type 2 diabetes &0.91&0.94&0.94&0.94&&0.94&0.94&0.94\\
 & Observational&0.60&0.89&0.83&0.85&&0.83&0.84&0.81 \\
\hline
HBV&Phase 3 cancer &0.93 &0.96&0.96&0.96&&0.96&0.98&0.96\\
 &Phase 1,2 cancer & 0.93&0.97&0.97&0.97&&0.96&0.98&0.97 \\
 & Heart disease&0.93&0.95&0.95&0.96&&0.92&0.89&0.98 \\
 &Type 2 diabetes &0.91&0.95&0.94&0.92&&0.91&0.93&0.96 \\
 &Observational& 0.74&0.88&0.88&0.90&&0.87&0.93&0.88 \\

 \hline
HCV&Phase 3 cancer &0.78&0.91&0.91&0.89&&0.70&0.89&0.82\\
 &Phase 1,2 cancer & 0.91&0.92&0.95&0.95&&0.92&0.92&0.92 \\
 & Heart disease&0.84&0.95&0.97&0.96&&0.82&0.92&0.92 \\
 &Type 2 diabetes &0.93&0.94&0.95&0.94&&0.96&0.95&0.95 \\
 &Observational& 0.61&0.81&0.87&0.85&&0.79&0.77&0.87\\

  \hline
Auto&Phase 3 cancer &0.98&0.99&0.99&0.99&&0.99&0.99&0.99\\
 &Phase 1,2 cancer & 0.91&0.93&0.93&0.93&&0.93&0.93&0.93 \\
 & Heart disease&0.91&0.83&0.85&0.85&&0.88&0.94&0.94\\
 &Type 2 diabetes &0.91&0.94&0.96&0.94&&0.94&0.94&0.94\\
 &Observational& 0.80&0.89&0.85&0.88&&0.79&0.87&0.89\\

\hline
Subst &Phase 3 cancer &0.12&0.96&0.96&0.96&&0.96&0.96&0.96\\
 &Phase 1,2 cancer &0.08&0.92&0.92&0.92&&0.92&0.92&0.92\\
 &Heart disease&0.34&0.93&0.93&0.93&&0.93&0.93&0.93 \\
 &Type 2 diabetes& 0.35&0.96&0.96&0.96&&0.96&0.96&0.96\\
 &Observational&0.39&0.96&0.96&0.96&&0.96&0.96&0.90\\
\hline
\end{tabular}
\caption{Few shot Criterion-level F1 Score for ClinicalTrialBERT. +\% refers to the percentage of keyword matches in our new dataset that are positive, which helps to contextualize the very high performance of some models.}
\label{table:results2}
\end{table*}

The result of few-shot learning is shown in Table~\ref{table:results2}, focusing just on criterion-level F1 scores for ClinicalTrialBERT.
We experimented with both the PROTECTOR1-trained model (the transfer learning scenario) and starting from the base ClinicalTrialBERT model (few-shot without transfer learning from the other dataset).
We selected random samples in increments of 5, 10, and 15 to fine-tune our dataset according to the relevant criteria and trial cohorts. And the sets of 5 and 10 are subsets of the larger set of 15 samples. This nested sampling approach ensures consistency in our fine-tuning process while allowing for a scalable analysis of the data.
The scores range from 0.42 (Prior cancer) to 0.99 (Auto).
Overall, the criterion-level model with few shot training performs better than the prior experiments, and in most cases, F1 scores improve with more training samples.
Specifically, 10-shot outperformed 5-shot in 34.3\% of cases, while 15-shot has the best average results, outperforming 10-shot in 35.7\% of cases and 5-shot in 42.9\% of cases.
When using the PROTECTOR1 data in addition to few-shot learning, this showed an improvement over omitting the PROTECTOR1 data in 22.9\% (5-shot), 34.3\% (10-shot), and 31.4\% (15-shot) of cases. 
Finally, using the best overall few-shot scenario (15-shot with PROTECTOR1 training), this outperformed the ClinicalTrialBERT criterion-level results from Table~\ref{table:results1} in 48.6\% of cases.


\section{Discussion}

The goal of this study was to evaluate the generalizability of eligibility classifiers for a diverse set of clinical trials.
Our starting point was the eligibility annotations in PROTECTOR1, which was limited to just phase 3 cancer trials. 
To accomplish the goal of generalizability evaluation, first and foremost a newly annotated dataset of clinical trial eligibility criteria was required.
We identified 5 cohorts of trials, ranging from phase 3 cancer trials just like PROTECTOR1 to observational trials that do not even have an intervention.
These trials were identified with high reliability using keyword heuristics, and further keyword heuristics were utilized to extract individual eligibility criteria statements for annotation.
The annotation process resulted in high levels of inter-annotator agreement.
The final annotated dataset contains 2,490 criteria from 500 clinical trials, relatively evenly distributed across the seven targeted eligibility exclusions.
This dataset will be made publicly available upon publication to encourage research in generalizable NLP models for eligibility classification.

As an initial evaluation of the generalizability of commonly-used BERT models, we experimented with two such models (ClinicalBERT and ClinicalTrialBERT) in both an out-of-the-box and few-shot learning scenario.
Our experimental results demonstrated that there was indeed a drop-off in performance going from phase 3 cancer trials to less-related trials.
The largest gap in performance in this respect was observational trials, which unlike the other four cohorts do not involve any kind of medical intervention and, as a result, their eligibility criteria can be substantially different.
Beyond this, the individual exclusion criteria that were disproportionately related to cancer trials (prior cancer history and HIV positivity) did worse than those that are more consistently seen across diseases.

Based on the result obtained through few-shot learning, the model's performance notably improves with the inclusion of a greater number of samples, especially when applied to cancer-specific trials. However, when considering different types of trials, additional training on original cancer data doesn't consistently offer a decisive advantage compared to not using this supplementary data.
Further, few-shot learning is not always applicable to many clinical trial-related tasks: while annotating a handful of disease-specific eligibility criteria is acceptable when performing a disease-specific task, any eligibility classification method that is disease-agnostic cannot afford even small numbers of annotations for every disease in ClinicalTrials.gov. Thus, both because few-shot learning does not always apply and because even its results are inferior to the in-domain results, more work on generalizabile models for eligibility classification is clearly needed.

To contextualize our results, which can often be quite high, we also provide results for a baseline approach entailing direct use of the keyword lists for each criterion, primarily with a focus on achieving high recall based on historical data. While this approach yielded favorable results for certain criteria, such as Auto, HBV, and HCV, it produced notably lower accuracy for other criteria. This highlights the importance of refining and curating a comprehensive yet precise set of keywords for each specific criteria, a vital step in enhancing the efficacy and accuracy of our model for future research endeavors.

It is also worth noting the challenges posed by data imbalance. Domain-specific models may tend to overfit to minority classes, as observed in previous research \cite{hsu2023investigation}. Future research efforts should prioritize the mitigation of data imbalance issues, especially for criteria that demonstrate pronounced prevalence in cancer trials. As previously mentioned, certain criteria, such as the exclusion of patients with prior malignancy and HIV positivity, are prevalent in cancer trials but typically do not apply to other cohorts. The substantial variation in training data becomes particularly salient when considering the application of cancer-specific fine-tuned models to diverse downstream tasks.

\section{Conclusion}
In conclusion, our research underscores the need for generalizable eligibility criteria classifiers. 
Our findings reveal the robust generalizability of these models to other phase 3 cancer trials and different phases within the realm of cancer studies. When extending the application to diseases that share similar eligibility requirements with cancer trials—such as those typically excluding HIV-positive and hepatitis B virus-positive participants—the cancer-trained models from cancer trials remain a strong and viable option. On the other hand, for more broadly defined trial types, exemplified by observational trials in our study, which don't specify a disease type and encompass diverse study objectives, the eligibility criteria can exhibit significant variations from trial to trial. 

Our few-shot learning results demonstrated significant enhancement when we introduced a few examples of previously unseen data. ClinicalTrialBERT initially achieved a criterion-level F1 score of only 0.69 for "phase 1,2 cancer trials," which subsequently improved to 0.94 after incorporating 10 additional samples into the model. Similarly, for "observational trials," ClinicalTrialBERT started with an F1 score of just 0.41 but made to 0.63 when we introduced 5 new samples to the model. While the few-shot learning approach yielded promising results, it wasn't always the case that performance improved with the addition of new samples. Several factors can contribute to this variation, including data variability. As the number of training shots increases, there can be more diversity within the training data, making it more challenging to discern consistent patterns. 
This underscores the complexity of training models to generalize from limited data and the need for a nuanced understanding of the data's inherent variability.

ClinicalTrialBERT, which was specifically trained on clinical trial eligibility criteria, continues to demonstrate its advantages 
compared to ClinicalBERT. 
Moreover, when applying the model to entirely new diseases or trials, it can be highly beneficial to incorporate a few previously unseen data samples into the model. Future research endeavors will encompass several key aspects. These include the clustering of criteria for various disease types and phases, addressing the class imbalance issue within the training data for specific diseases, and assessing the model's applicability to additional cancer-specific criteria.\\
The future direction entails concentrating on a specific array of diseases. For instance, even within cancer trials, differences arise based on various cancer types, stages, and treatment modalities. Addressing the question posed when considering alternative diseases like A and/or B, diseases deemed "similar" from a trial eligibility standpoint should be grouped together, followed by conducting additional assessments for these clustered diseases.

\section{References}\label{sec:reference}

\bibliographystyle{lrec-coling2024-natbib}
\bibliography{lrec-coling2024-example}

\end{document}